\documentclass[runningheads]{llncs}
\usepackage[T1]{fontenc}

\usepackage{amsmath}
\usepackage{amsfonts}
\usepackage{algorithm}
\usepackage{algpseudocode}
\usepackage{multicol}
\usepackage{graphicx}
\usepackage{multirow}
\usepackage{subcaption}

\usepackage{algpseudocode}
\usepackage{stmaryrd}

\usepackage{breakcites}
\usepackage{booktabs}
\usepackage{hyperref}
\usepackage{color}

\hypersetup{
    colorlinks,
    linkcolor={blue},
    citecolor={blue},
    urlcolor={blue}
}

\usepackage{cleveref}

\newcommand{\coloneqq}{\mathrel{\mathop:}=}

\begin{document}

\title{Spherical Brownian Bridge Diffusion Models for Conditional Cortical Thickness Forecasting}
\titlerunning{Spherical Brownian Bridge Diffusion Model}

\author{%
  Ivan Stoyanov\inst{1*} \and
  Fabian Bongratz\inst{1,2*}\and
  Christian Wachinger\inst{1,2}
}
\authorrunning{I. Stoyanov, F. Bongratz, C. Wachinger}
\institute{Lab for AI in Medical Imaging, Technical University of Munich, Munich, Germany 
\\
\email{stoyaniv@in.tum.de, fabi.bongratz@tum.de}
\and
Munich Center for Machine Learning, Munich, Germany
\\
$^*$Equal contribution
}

 \maketitle

\begin{abstract}
Accurate forecasting of individualized, high-resolution cortical thickness (CTh) trajectories is essential for detecting subtle cortical changes, providing invaluable insights into neurodegenerative processes
and facilitating earlier and more precise intervention strategies.
However, CTh forecasting is a challenging task due to the intricate non-Euclidean geometry of the cerebral cortex and the need to integrate multi-modal data for subject-specific predictions. 
To address these challenges, we introduce the Spherical Brownian Bridge Diffusion Model (SBDM). Specifically, we propose a bidirectional conditional Brownian bridge diffusion process to forecast CTh trajectories at the vertex level of registered cortical surfaces.
Our technical contribution includes a new denoising model, the conditional spherical U-Net (CoS-UNet), which combines spherical convolutions and dense cross-attention to integrate cortical surfaces and tabular conditions seamlessly.
Compared to previous approaches, SBDM achieves significantly reduced prediction errors, as demonstrated by our experiments based on longitudinal datasets from the ADNI and OASIS. 
Additionally, we demonstrate SBDM's ability to generate individual factual and counterfactual CTh trajectories, offering a novel framework for exploring hypothetical scenarios of cortical development.

\end{abstract}

\section{Introduction}

\begin{figure}[t]
    \centering
    \includegraphics[width=0.9\linewidth]{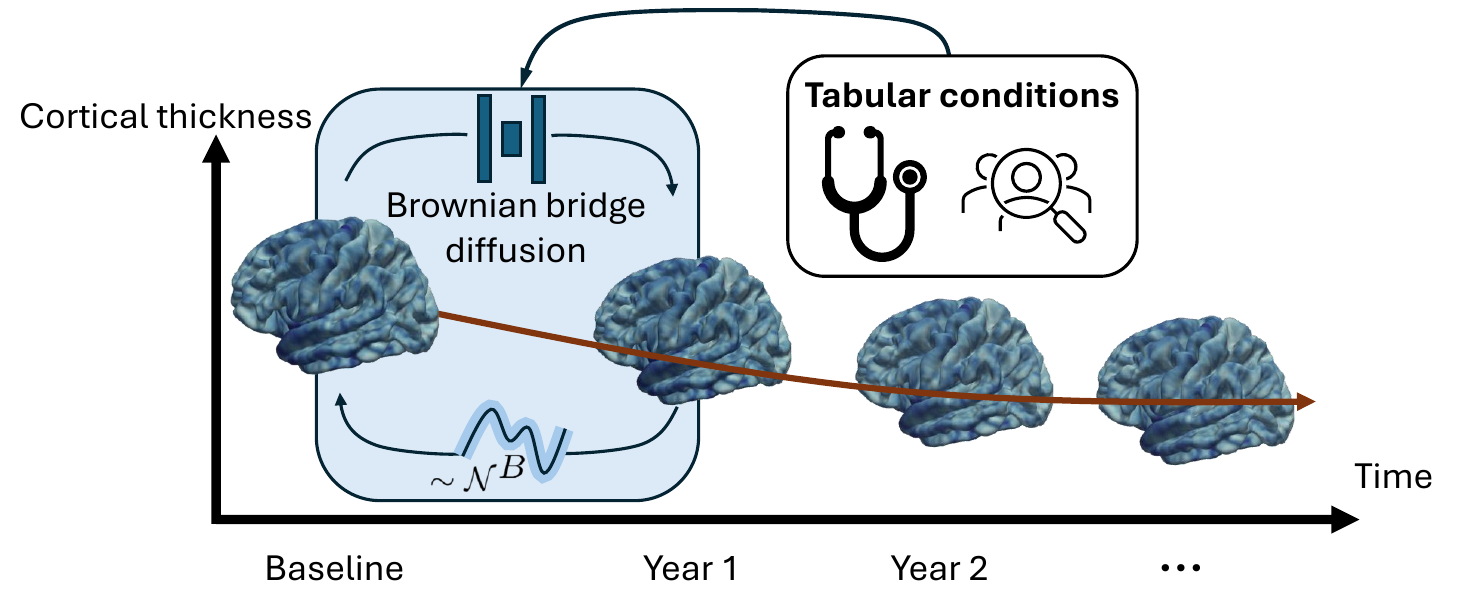}
    \caption{Illustration of cortical thickness~(CTh) trajectories across time. Based on an individual's baseline data, including observed CTh and tabular data such as demographics and disease conditions, we aim to predict future CTh progression. 
    }
    \label{fig:overview}
\end{figure}

Progressive loss of cortical gray matter is a hallmark of neurodegenerative diseases such as Alzheimer’s disease (AD) \cite{Scahill2002atrophyevolutionAD}, establishing cortical thickness (CTh) as a critical biomarker for tracking disease progression \cite{Risacher2010longitudinalatrophy,schwarz2016,Singh2006corticalthinningalzheimers}. 
The reconstruction of cortical surfaces from magnetic resonance imaging (MRI) enables subvoxel-accurate CTh measurements at the vertex level, providing detailed insights into the structure of the cerebral cortex~\cite{fischl2000}. 
However, single MRI scans only capture a static snapshot of the dynamic aging and disease processes. 
\emph{Forecasting} future changes in CTh from a single scan, as illustrated in \Cref{fig:overview}, could offer invaluable insights into disease progression, enabling early diagnosis, optimized clinical trial design, and personalized treatment strategies.
\emph{Vertex-wise} prediction of highly detailed CTh maps is particularly relevant, as it can capture subtle, localized changes to distinguish between normal age-related cortical thinning and pathological alterations. 
However, accurate forecasting is inherently challenging due to (i) the complex non-Euclidean geometry of the cortical sheet, (ii) the need to integrate multi-modal data for subject-specific predictions, and (iii) the irregularity of longitudinal datasets. 

To address these challenges, we introduce the \underline{S}pherical Brownian \underline{B}ridge \underline{D}iffusion \underline{M}odel (SBDM). Grounded in seminal work on 2D image-to-image translation~\cite{li_bbdm_2023}, we propose to model CTh trajectories with a bi-directional Brownian bridge diffusion process, as illustrated in \Cref{fig:overview}. SBDM builds on diffusion models, which have shown remarkable success in generating realistic images in both natural and medical imaging domains. Unlike the commonly used denoising diffusion probabilistic models (DDPMs) \cite{ddpm,sohl-dickstein-diffusion2015,xiao2024miccai,xie2024cortical}, we leverage the recently proposed Brownian bridge diffusion model (BBDM)\cite{li_bbdm_2023}, originally developed for 2D natural images. BBDM stochastically maps between structured start- and endpoints, preserving data-specific characteristics while avoiding the introduction of pure Gaussian noise.
Thanks to the bidirectional diffusion of BBDM, we can set the high-dimensional CTh maps as input, while low-dimensional tabular data (demographics, time, and diagnoses) are incorporated as conditions. 
Unlike DDPM, which treats both data types as conditions and requires balancing their different dimensionalities, BBDM inherently accommodates this multi-modal integration. 
To handle the intricate geometry of the cortical surface, we developed a dedicated conditional spherical U-Net (CoS-UNet). 
This architecture maps cortical thickness data onto a spherical representation, employing spherical convolutions to capture local spatial relationships, and integrating cross-attention layers to model global dependencies from tabular conditional variables. 
Our experiments demonstrate the effectiveness of SBDM, significantly outperforming existing methods for vertex-wise prediction. 
We further demonstrate SBDM's capability to generate individual factual and counterfactual trajectories by conditioning on a target diagnosis.

\section{Related Work}
Prior spatiotemporal models capable of vertex-wise prediction of structural brain development often aggregate vertex-wise measures on a region level~\cite{perez-millan2024,xiao2024miccai}, relying on specific brain atlases and suffering from substantial loss of spatial detail~\cite{Fuertjes2023}.
A recent DDPM for region-level CTh forecasting (CTh-DDPM)~\cite{xiao2024miccai} shows promise but fails to generalize to vertex-level predictions, as demonstrated in our experiments (\Cref{sec:accuracy}). 
Spherical U-Net~\cite{zhao_spherical_2019} and Spherical Transformers~\cite{Cheng2022spherical,dahan2022sit} provide vertex-level predictions, but they do not incorporate tabular conditional variables. Hence, they are restricted to predicting fixed time intervals under constant conditions, limiting their applicability to real-world longitudinal data~\cite{Cheng2022spherical,zhao_spherical_2019}. 
Finally, parametric models~\cite{Marinescu2019dive,Young2024diseaseprogressionreview} can extrapolate longitudinal data on the vertex level but require observations from multiple visits and impose simplistic assumptions on the shape of trajectories.

\section{Methods}
In this section, we describe the \underline{S}pherical Brownian \underline{B}ridge \underline{D}iffusion \underline{M}odel (SBDM) for forecasting vertex-wise longitudinal cortical thickness trajectories. We explain the Brownian bridge diffusion process on cortical surfaces in \Cref{sec:cortex-diffusion} and the accompanying denoising network in \Cref{sec:denoising}. Throughout this work, we consider longitudinal data from $N$ subjects, $1\leq i \leq N$, and an arbitrary number of $M_i\geq1$ follow-up visits per subject. Each subject $i$ has a baseline scan with associated cortical thickness, $\tau_0^{(i)}\in \mathbb{R}^{|V|}$, and a set of follow-up measurements $\{\tau_{t_j}^{(i)} \mid \tau_{t_j}^{(i)} \in \mathbb{R}^{|V|} \wedge 1 \leq j \leq M_i\}$, resulting in $M_i+1$ timepoints per subject. For the computation of cortical thickness~(CTh) from T1w MRI scans, we used longitudinal FreeSurfer~\cite{fischlFreeSurfer2012,Reuter2012longfreesurfer}. We registered and resampled all data to FreeSurfer's FsAverage brain template, obtaining a fixed number of $|V|$ CTh values for each timepoint.

\subsection{Spherical Brownian Bridge Diffusion on Cortical Surfaces}\label{sec:cortex-diffusion}

\begin{figure}[t]

    \centering
    \includegraphics[width=\textwidth]{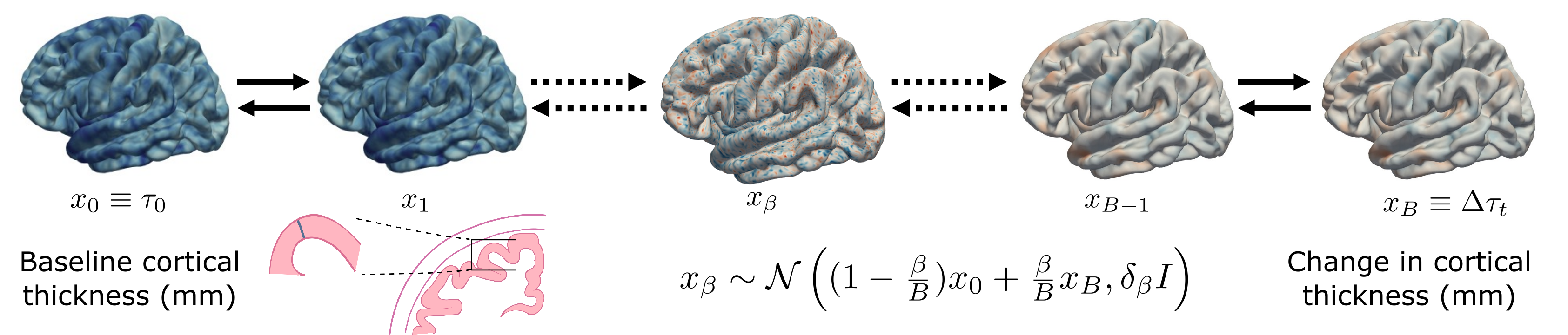}
    \caption{
    Illustration of the proposed Spherical Brownian Bridge Diffusion Model~(SBDM) on cortical surfaces. SBDM maps between current (baseline) cortical thickness $\tau_0$  and a future relative change $\Delta \tau_t$. 
    }
    \label{fig:bbdm-cth}
\end{figure}

A Brownian bridge is a continuous stochastic process that has recently been introduced for image-to-image translation~\cite{li_bbdm_2023}. The core idea of the Brownian bridge diffusion model (BBDM) is to map stochastically between specific start- and endpoints, $x_0$ and $x_B$, while preserving structural characteristics:
\begin{equation} \label{eq:bridge}
    q(x_\beta \mid x_0, x_B) = \mathcal{N}\left(x_\beta; \, (1 - m_\beta) x_0 + m_\beta x_B, \delta_\beta I \right),
\end{equation}
where $0\leq\beta\leq B$ and $m_\beta\coloneqq\frac{\beta}{B}\in [0,1]$. The variance schedule is defined as $\delta_\beta\coloneqq 2(m_\beta-m_\beta^2)$, reaching its maximum value of $\delta_\beta=1/2$ at the midpoint of the bridge, i.e., for $m_\beta = 0.5$. 
In contrast to the standard Brownian bridge, where the forward process maps \(x_0 \mapsto x_B\), our formulation explicitly reverses this direction: by definition, our forward process maps the relative change \(\Delta \tau_t\) at some arbitrary future timepoint $t$ (corresponding to \(x_B\)) and maps it to the baseline thickness \(\tau_0\) (corresponding to \(x_0\)). Conversely, the reverse process reconstructs \(\Delta \tau_t\) from \(\tau_0\). 
We illustrate this process in \Cref{fig:bbdm-cth}. Note, the step $\beta$ of the stochastic process is not equivalent to the physical time $t$; its main purpose is to permit an iterative metamorphosis.
Formally, the relation between baseline CTh $\tau_0$, relative change $\Delta \tau_t$, and follow-up CTh $\tau_t$ is given by
\begin{align} \label{eq:cth}
    \tau_t &\equiv \tau_0 + \Delta \tau_t.
\end{align}

\noindent
\textbf{Conditioning.}
The baseline cortical thickness provides highly individualized information for the prediction of future brain development. However, further demographic and clinical factors need to be considered for accurate modeling of longitudinal brain morphology~\cite{BernalRusiel2013linearmixedmodels}. To this end, we propose to incorporate age $a$, sex $s$, baseline diagnosis $d_0$, and follow-up diagnosis $d_t$\footnote{Follow-up diagnosis is only added for the counterfactual CTh experiment (Sec. \ref{sec:trajectories}).} into SBDM. For the sake of compactness, we summarize these covariates in the conditional variable $c=(a, s, d_0, d_t)$. Finally, we add the time $t$ between the baseline and the follow-up visit to our model to tailor the prediction to a specific time interval. In contrast to previous works~\cite{Chai2021vaechanges,Sarasua2022cashformer,zhao_spherical_2019}, conditioning explicitly on the time $t$ makes the model agnostic to incomplete study data and provides flexibility in the modeled trajectories.

\noindent
\textbf{Training.}
We outline the training cycle of SBDM in \Cref{alg:training_bbdm}. Given longitudinal training data, i.e., baseline cortical thickness $\tau_0$ and a relative change $\Delta \tau_t$, we sample intermediate states $x_\beta$ based on \Cref{eq:bridge}. Then, we train a neural network $f_\theta$ to recover the ``noise'' from $x_\beta$, which amounts to minimizing the following loss function:
\begin{equation} \label{eq:loss}
\mathcal{L}_{\text{SBDM}} = \left\lVert (1 - m_\beta) (\tau_0 - \Delta \tau_t) + \sqrt{\delta_\beta} \epsilon - f_\theta(x_\beta, \beta, t, c) \right\rVert^2.
\end{equation}
The noise $\epsilon$ is sampled from a standard multivariate normal distribution, i.e., $\epsilon \sim \mathcal{N}(\mathbf{0}, \mathbf{I})
$. 
However, different from previous Brownian bridge-based models~\cite{Lee2024ebdm,li_bbdm_2023}, we operate directly in the space of cortical thickness maps, i.e., $\epsilon \in \mathbb{R}^{|V|}$, instead of an abstract autoencoder latent space. This makes our model more interpretable and removes the dependency on potentially suboptimal decoding from the latent space, which can hinder high-resolution predictions and degrade the quality of fine-grained outputs~\cite{Rombach2022}.

\begin{algorithm}[t]
\caption{SBDM Training}\label{alg:training_bbdm}
\hspace*{\algorithmicindent} \textbf{Input:} Longitudinal training set $\mathcal{D}_{\text{train}}$\\
\hspace*{\algorithmicindent} \textbf{Output:} Trained model $f_\theta$
\begin{algorithmic}[1]
\Repeat
    \State $(\tau_0, \Delta \tau_t, t, c) \sim \mathcal{D}_{\text{train}}$
    \Comment Get training sample
    \State $\beta \sim \text{Uniform}(0, \dots, B)$
    \Comment Sample step in the bridge
    \State $\epsilon \sim \mathcal{N}(\mathbf{0}, \mathbf{I})$
    \Comment Sample Gaussian noise
    \State $\delta_\beta \gets 2(m_\beta-m_\beta^2)$
    \Comment Get variance
    \State $x_\beta \gets (1 - m_\beta)\tau_0 + m_\beta \Delta \tau_t + \sqrt{\delta_\beta} \epsilon$
    \Comment Intermediate state in bridge
    \State Take gradient descent step on
    $
    \nabla_\theta \| (1 -m_\beta) (\tau_0-\Delta \tau_t) + \sqrt{\delta_\beta} \epsilon - f_\theta(x_\beta, \beta, t, c) \|^2
    $
\Until{converged}
\State \Return $f_\theta$
\end{algorithmic}
\end{algorithm}

\noindent
\textbf{Inference.}
\Cref{alg:inference_bbdm} describes the inference with a trained SBDM. 
According to \Cref{eq:bridge}, we can sample intermediate representations $x_\beta$ if both endpoints of the bridge, i.e., $\tau_0$ and $\Delta \tau_t$, are available. While this is the case at training time, our aim is to predict $\Delta \tau_t$ at inference time, and it is therefore not available. Extending the iterative stochastic process from~\cite{li_bbdm_2023} with additional covariates $c$ and time difference $t$, we employ the following recursion to arrive at $x_B\equiv\Delta \tau_t$:
\begin{equation} \label{eq:inference}
    p_\theta\left(x_{\beta + 1} \mid x_\beta, \tau_0, t, c \right) = \mathcal{N}\left(x_{\beta+1}; \, \mu_\theta(x_\beta, \beta, t, c), \tilde{\delta}_\beta I\right).
\end{equation}
Unlike DDPM-like diffusion models, which start the inference process from Gaussian noise and incorporate conditions through the denoising model, our Brownian bridge-based approach engraves the most critical condition for prediction, the baseline CTh, directly into the starting point of the generation process.
Additional covariates such as age, sex, and diagnosis further guide the process for best accuracy. 

We sample $x_{\beta+1}$ with probability in \Cref{eq:inference} via the recusion
\begin{equation}
    x_{\beta+1} = \zeta_{1, \beta} x_\beta + \zeta_{2, \beta} \tau_0 - \zeta_{3, \beta}f_\theta(x_\beta, \beta, t, c) + \sqrt{\tilde{\delta}_\beta} \eta
\end{equation}
in analogy to the original BBDM~\cite{li_bbdm_2023}. We provide details about the coefficients $\zeta_{1, \beta}$, $\zeta_{2, \beta}$, $\zeta_{3, \beta}$, and $\tilde{\delta}_\beta$, in \Cref{alg:inference_bbdm}. 
Finally, we use a non-Markovian sampling strategy~\cite{li_bbdm_2023,ddim} in SBDM to speed up the inference process. Instead of traversing through all $B$ stages of the Brownian bridge, we predict only a subset $\{x_{\bar{\beta}_1}, \ldots, x_{\bar{\beta}_{\bar{B}}}\}$ of $\{x_1, \ldots,$ $x_B\}$.

\begin{algorithm}[t]
\caption{SBDM Inference} \label{alg:inference_bbdm}
\hspace*{\algorithmicindent} \textbf{Input:} Baseline thickness $\tau_0$, time difference $t$, and covariates $c$\\
\hspace*{\algorithmicindent} \textbf{Output:} Cortical thickness change $\Delta \tau_t$
\begin{algorithmic}[1]

\State $x_0 \gets \tau_0$
\Comment Set starting point
\For{$\beta = 0, \dots, B-1$}
\Comment Iterate over bridge
    \State $\eta \sim \mathcal{N}(\mathbf{0}, \mathbf{I})$
    \Comment Sample Gaussian noise
    \State $\delta_{\beta|\beta+1} \gets \delta_\beta - \delta_{\beta+1}\frac{m_\beta^2}{m_{\beta+1}^2}$
    \State $\zeta_{1, \beta} \gets \frac{\delta_{\beta+1}}{\delta_\beta}\frac{m_\beta}{m_{\beta+1}} + \frac{\delta_{\beta|\beta+1}}{\delta_\beta}m_{\beta+1}$
    \State $\zeta_{2, \beta} \gets 1 - m_{\beta+1} - (1 - m_\beta) \frac{m_\beta}{m_{\beta+1}} \frac{\delta_{\beta+1}}{\delta_\beta}$
    \State $\zeta_{3, \beta} \gets m_{\beta+1}\frac{\delta_{\beta|\beta+1}}{\delta_\beta}$, $\tilde{\delta}_\beta \gets \delta_{\beta|\beta+1} \cdot \delta_{\beta+1}/{\delta_\beta}$
    \Comment Get coefficients
    \State $x_{\beta+1} \gets \zeta_{1, \beta} x_\beta + \zeta_{2, \beta} \tau_0 - \zeta_{3, \beta}f_\theta(x_\beta, \beta, t, c) + \sqrt{\tilde{\delta}_\beta} \eta$ 
    \Comment Compute recursion
\EndFor
\State $\Delta \tau_t \gets x_B$ 
\Comment Get final value
\State \Return $\Delta \tau_t$
\end{algorithmic}
\end{algorithm}

\begin{figure}[t]
    \centering
    \includegraphics[width=\textwidth]{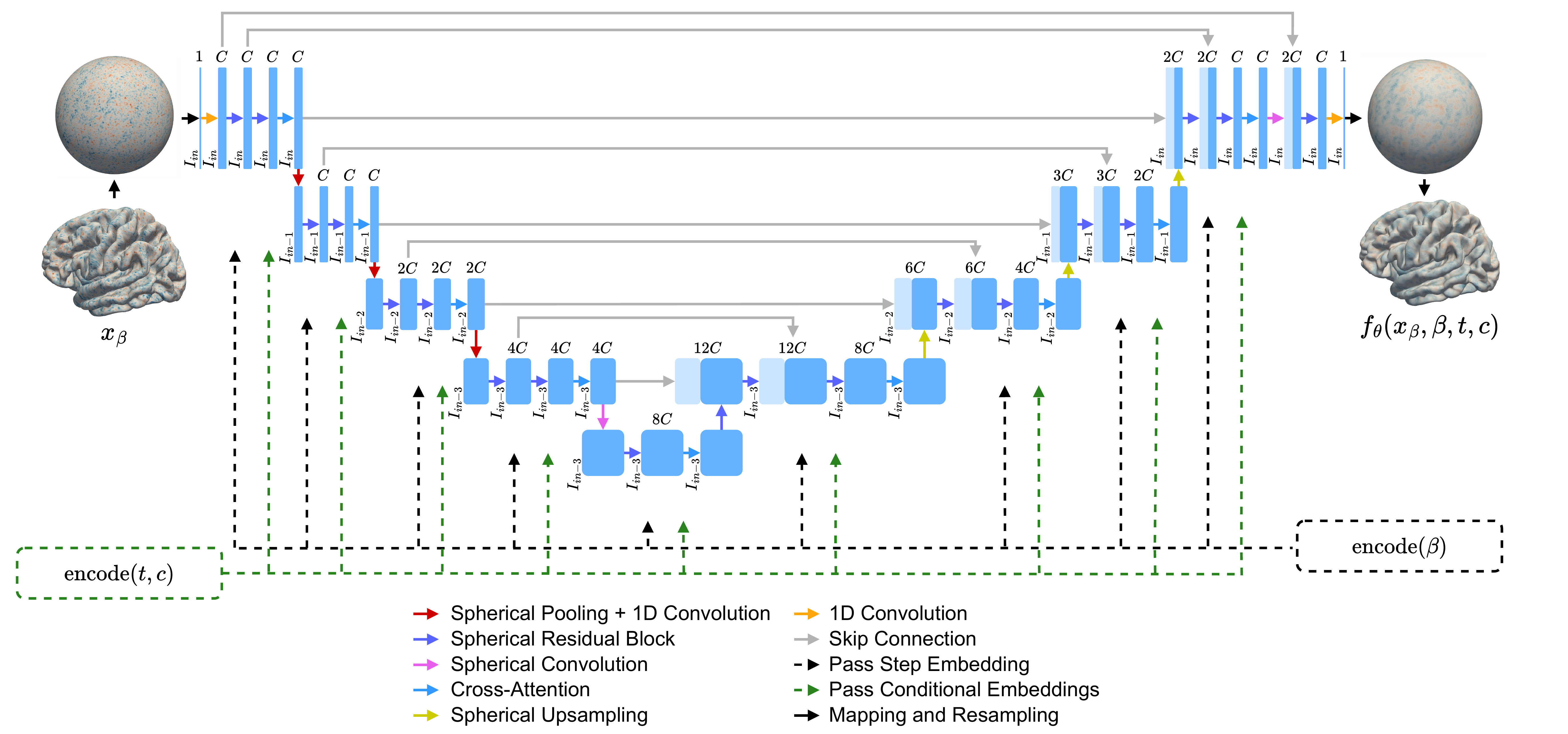}
    \caption{
    Illustration of the proposed Conditional Spherical U-Net (CoS-UNet).
    }
    \label{fig:ssdnet}
\end{figure}

\subsection{Conditional Spherical U-Net (CoS-UNet)} \label{sec:denoising}
This section describes the architecture of the denoising neural network $f_\theta$, which is a central component in SBDM. 
Unlike prior work that employed the Brownian bridge for image data~\cite{Lee2024ebdm,li_bbdm_2023}, SBDM operates on triangular surface meshes. 
Specifically, cortical thickness measurements are mapped to geodesic polyhedra, also called \emph{icospheres} since they are obtained through recursive subdivision of an icosahedron. 
To operate in this domain, we build our denoising network on the basis of Spherical UNet~(S-UNet)~\cite{zhao_spherical_2019}, a model designed for spherical cortical morphology data. 
Additionally, we explored other reasonable architectures like MLPs and surface vision transformers~(SiT)~\cite{dahan2022sit}; see \Cref{sec:ablation} for a detailed comparison.

The original Spherical U-Net does not natively accept conditioning variables.  
We therefore add three inputs required by the denoising formulation: the Brownian–bridge step $\beta$, the follow-up interval $t$, and covariates $c$.  
The time interval $t$ and the covariates $c$ are embedded jointly and injected into every residual stage through a cross-attention layer, giving the network dense, stage-wise guidance.  
The step index $\beta$ is embedded separately and added to the latent feature maps, so that $\beta$ remains disentangled from $(t,c)$.

The Conditional Spherical U-Net (CoS-UNet), shown in \Cref{fig:ssdnet}, adopts a classical encoder–decoder layout~\cite{Ronneberger2015unet} with residual links and spherical pooling that inverts the icosphere subdivision after each down-sampling step.  
Every encoder stage contains two spherical residual blocks; each block consists of  
(i) a one-hop spherical graph convolution~\cite{zhao_spherical_2019},  
(ii) group normalization~\cite{yuxin2018groupnorm}, and  
(iii) a SiLU activation layer~\cite{Elfwing2018silu}.  
A cross-attention layer follows the second residual block to integrate the conditioning information described above.
The decoder mirrors the encoder: transposed spherical convolutions up-sample the feature maps step by step until the initial icosphere resolution is restored, while skip connections preserve high-resolution details.

\section{Results and Discussion}

\subsection{Experimental Setup}

\noindent
\textbf{Data.}
We utilized data from the Alzheimer's Disease Neuroimaging Initiative (ADNI, \url{https://adni.loni.usc.edu}). The database contains longitudinal visits from cognitively normal (CN) subjects, subjects with mild cognitive impairment~(MCI), and subjects diagnosed with Alzheimer's disease (AD). The maximum time period is 168 months. At each visit, the diagnosis was made based on clinical evaluations, cognitive tests, and biomarker measurements\footnote{\url{https://adni.loni.usc.edu/wp-content/uploads/2012/08/instruction-about-data.pdf}}. We split the data at the subject level, accounting for age (55--97 years), sex, diagnosis, and number of follow-up visits. This resulted in 921 subjects (4,112 scans) for training, 306 subjects (1,412 scans) for validation, and 306 subjects (1,387 scans) for testing. We extracted cortical thickness maps with the longitudinal FreeSurfer~\cite{fischlFreeSurfer2012} stream~(v7.2), which reduces the variability in the measurements by estimating unbiased templates for each subject~\cite{Reuter2012longfreesurfer}. Afterward, we registered cortical surfaces to the FsAverage icosphere template~\cite{Fischl1999fsaverage}.
To assess the generalization of our trained model, we further used data from the Open Access Series of Imaging Studies (OASIS, \url{https://sites.wustl.edu/oasisbrains}) for testing only. We considered 1,750 scans from 590 subjects (age 42--95 years), distinguishing CN and AD cases. We ignored the medial region that connects the two brain hemispheres unless stated otherwise.

\noindent
\textbf{Implementation Details.}
We implemented SBDM based on PyTorch (v2.0.1), Cuda~(v12.2), and the Sphericalunet package~(v1.2.2), using an implementation of DDPM~\cite{ddpm} as a starting point\footnote{\url{https://github.com/lucidrains/denoising-diffusion-pytorch}}. 
All models were trained on a single Nvidia Titan RTX GPU with 24\,GB of VRAM.
Our source code will be made available online at \url{https://github.com/ai-med/SBDM}.
With the number of channels in CoS-UNet set to $C=64$, we trained the SBDM for a maximum of 2,000 epochs, selecting the best model based on the validation set. During training, we set the horizon of the Brownian bridge to $B=1,000$. At inference, we sample 200 intermediate stages in the bridge as suggested in previous studies to speed up the sampling process~\cite{li_bbdm_2023,ddim}. 
The AdamW optimizer was employed with an initial learning rate of $1\mathrm{e}{-4}$, which we reduced on plateaus after 100 epochs without validation loss improvement. 
Additionally, we applied an exponential moving average to the training weights to obtain the final model.

\noindent
\textbf{Reference Methods.}
To benchmark the performance of SBDM, we adapted several existing methods for vertex-wise prediction of cortical morphology. These include a linear regression model, Spherical U-Net~\cite{zhao_spherical_2019}, Surface Vision Transformer~(SiT)~\cite{dahan2022sit}, and a recent conditional diffusion model explicitly designed for cortical thickness prediction~(CTh-DDPM)~\cite{xiao2024miccai}. 
Given its conceptual similarity to SBDM, we re-implemented CTh-DDPM to ensure optimal comparability.
Originally, CTh-DDPM uses a 1D denoising network that does not capture spatial relationships across the cortex. Hence, we also implemented a version using the CoS-UNet as the denoising network within DDPM~(DDPM/CoS-UNet).
We adapted the Spherical U-Net and CTh-DDPM to incorporate the same conditional input variables as SBDM~(in a similar manner as in CoS-UNet, i.e., by addition to latent features) and trained them consistently on our ADNI training set. For testing on OASIS, we used all models without fine-tuning.

\begin{table}[t]
    \renewcommand\bfdefault{b}
    \centering
    \setlength{\tabcolsep}{3pt}
    \caption{Mean absolute error (mean\textpm SD) in mm of predicted CTh based on the ADNI test set and OASIS data. Values were aggregated across the entire cohort (all) and separate for each diagnostic subgroup.
    }
    \label{tab:combined_evaluation}
    \begin{tabular}{l l cccc}
        \toprule
        Dataset & Method & All & CN & MCI & AD \\
        \midrule
        \multirow{4}{*}{ADNI} 
        & Linear regression & 0.135\textpm0.031 & 0.130\textpm0.030 & 0.135\textpm0.032 & 0.145\textpm0.027 \\
        & Spherical U-Net~\cite{zhao_spherical_2019} & 0.111\textpm0.029 & 0.106\textpm0.022 & 0.110\textpm0.031 & 0.126\textpm0.031 \\
        & SiT~\cite{dahan2022sit} & 0.108\textpm0.028 & 0.103\textpm0.021 & 0.106\textpm0.030 & 0.120\textpm0.030 \\
        & CTh-DDPM~\cite{xiao2024miccai} & 0.208\textpm0.019 & 0.207\textpm0.015 & 0.208\textpm0.023 & 0.211\textpm0.019 \\
        & DDPM/CoS-UNet & 0.119\textpm0.029 & 0.115\textpm0.024 & 0.117\textpm0.032 & 0.130\textpm0.030 \\
        & SBDM & \textbf{0.097\textpm0.031} & \textbf{0.092\textpm0.024} & \textbf{0.094\textpm0.033} & \textbf{0.112\textpm0.034} \\
        \midrule
        \multirow{4}{*}{OASIS} 
        & Linear Regression & 0.146\textpm0.154 & 0.146\textpm0.157 & -- & 0.146\textpm0.048 \\
        & Spherical U-Net~\cite{zhao_spherical_2019} & 0.115\textpm0.027 & 0.115\textpm0.027 & -- & 0.120\textpm0.034 \\
        & SiT~\cite{dahan2022sit} & 0.112\textpm0.025 & 0.112\textpm0.025 & -- & 0.119\textpm0.033 \\
        & CTh-DDPM~\cite{xiao2024miccai} & 0.210\textpm0.017 & 0.210\textpm0.017 & -- & 0.212\textpm0.019 \\
        & DDPM/CoS-UNet & 0.121\textpm0.030 & 0.121\textpm0.030 & -- & 0.124\textpm0.033 \\
        & SBDM & \textbf{0.100\textpm0.030} & \textbf{0.099\textpm0.030} & -- & \textbf{0.105\textpm0.035} \\
        \bottomrule
    \end{tabular}%
\end{table}

\begin{figure}[t]
    \centering
    \includegraphics[width=\linewidth]{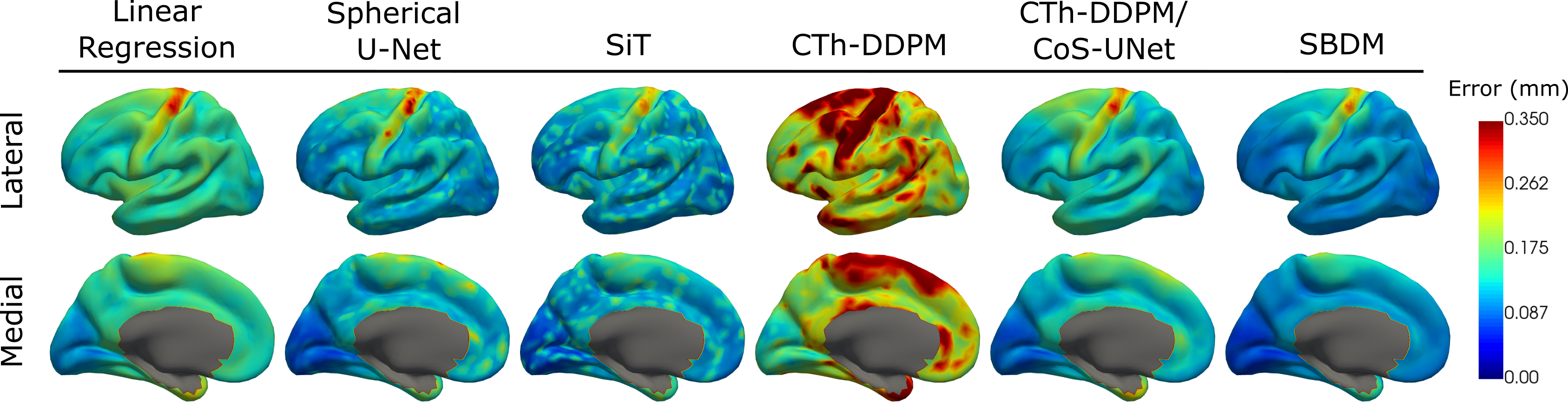}
    \caption{Mean vertex-level prediction errors based on the ADNI test set.}
    \label{fig:vertex-plot}
\end{figure}
\subsection{Forecasting Accuracy} \label{sec:accuracy}

We report the accuracy in terms of mean absolute error~(MAE) of predicted vertex-wise CTh on the left brain hemisphere in \Cref{tab:combined_evaluation}. Additionally, we depict vertex-wise errors on the FsAverage template in \Cref{fig:vertex-plot}.

Generally, the deep learning-based methods outperform the basic linear regression for CTh prediction considerably, with the exception of the CTh-DDPM model. However, with our CoS-UNet, the DDPM closes the gap to the other methods, indicating that the limitation in the original CTh-DDPM likely results from its 1D U-Net, which was designed for region-level data~(34 DKT regions instead of thousands of vertices). The SiT and the Spherical U-Net provide strong baselines, leveraging both the conditioning variables and the baseline CTh features to predict future changes in CTh. Yet, the proposed SBDM excels by yielding the best results across all diagnostic groups and datasets. It outperforms the second-best method, SiT, by around 0.01\,mm, which corresponds to an improvement of about 10\%.
Among the diagnostic subgroups, the CN group achieved the lowest errors, with the SBDM model reaching an MAE of 0.092\,mm on the ADNI dataset. For MCI and AD subgroups, the errors are slightly higher (0.094\,mm and 0.112\,mm, respectively), which is in accordance with previous related work~\cite{xiao2024miccai}. This could be due to the heterogeneity of atrophy patterns in Alzheimer's disease~\cite{Poulakis2018heterogeneous}, thereby being more difficult to predict than the healthy aging process. 

When transferred to the OASIS study, of which no scans were seen during training, SBDM remains at the forefront of all implemented methods with an average MAE of 0.100\,mm. Compared to the MAE on the ADNI test set, the error increases only slightly for the CN subgroup (+0.003\,mm); for the AD group, the error is even lower on external OASIS data (-0.007\,mm). This demonstrates the good generalization of SBDM and its applicability to new datasets without retraining.
Paired, two-sided Wilcoxon signed-rank tests revealed statistically significant improvements of SBDM compared to the four pre-existing reference methods on both datasets ($p<10^{-3}$).

Complementing the findings from \Cref{tab:combined_evaluation}, the vertex-wise errors in \Cref{fig:vertex-plot} reveal that the improvement of SBDM over the other methods is consistent across the entire cortical sheet. Interestingly, the highest errors occurred for all methods in the precentral gyrus, an area that is usually less affected by cortical atrophy in AD~\cite{Singh2006corticalthinningalzheimers}. Still, SBDM yields comparably low errors in this area and avoids dotted heterogeneities in Spherical U-Net and SiT.
\begin{table}[t]
    \setlength{\tabcolsep}{3pt}
    \renewcommand\bfdefault{b}
    \centering
    \caption{Ablation of the denoising model in SBDM based on the ADNI validation set; reporting as in \Cref{tab:combined_evaluation}.}
    \label{tab:ablation}
    \begin{tabular}{l c cccc}
        \toprule
        Denoising Model & All & CN & MCI & AD \\
        \midrule
        Spherical U-Net-based & 0.281\textpm0.027 & 0.278\textpm0.022 & 0.277\textpm0.028 & 0.294\textpm0.028\\
        MLP-based & 0.269\textpm0.024 & 0.263\textpm0.019 & 0.267\textpm0.024 & 0.282\textpm0.026\\
        SiT-based & 0.097\textpm0.029 & 0.094\textpm0.025 & 0.093\textpm0.025 & 0.112\textpm0.037 \\
        CoS-UNet & \textbf{0.095\textpm0.029} & \textbf{0.092\textpm0.025} & \textbf{0.092\textpm0.025} & \textbf{0.109\textpm0.037} \\        
        \bottomrule
    \end{tabular}%
\end{table}

\subsection{Ablation Study of the Denoising Model} \label{sec:ablation}
In \Cref{tab:ablation}, we assess the impact of the proposed CoS-UNet on the performance of SBDM
based on our ADNI validation set. Specifically, we report results from SBDM variants where we replaced CoS-UNet with reasonable alternatives, i.e., with neural networks~(NNs) that can operate on registered spherical data. To this end, we implemented a fully-connected multi-layer perceptron~(MLP), the original Spherical U-Net~\cite{zhao_spherical_2019}, a 1D U-Net as used in CTh-DDPM~\cite{xiao2024miccai}, and the Surface Vision Transformer~(SiT)~\cite{dahan2022sit}. We implemented all NNs consistently into our SBDM framework and adapted them to take the same input, i.e., baseline cortical thickness and conditions, as our CoS-UNet. 

We find that CoS-UNet yields the highest prediction accuracy for SBDM, closely followed by the SiT. The other NNs~(MLP, Spherical U-Net, and 1D U-Net) are not competitive with these architectures, speaking to the benefit of the attention mechanism. Nevertheless, the CoS-UNet maintains an edge in accuracy over the SiT across all diagnostic groups, making it the best choice for SBDM.

\subsection{Individual Conditional Disease Trajectories}\label{sec:trajectories}
\begin{figure}[t]
    \centering

    \includegraphics[width=\textwidth]{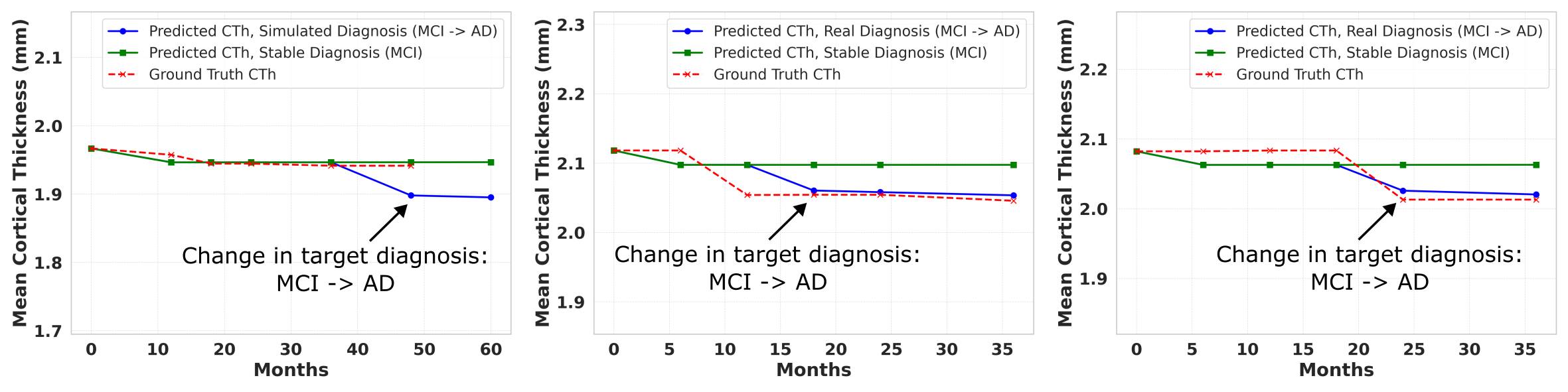}

    \caption{Simulated CTh trajectories for three individuals with varying target diagnoses. From left to right: 67 years, male; 71 years, male; 77 years, female.}
    \label{fig:trajectories}
\end{figure}

So far, we have presented results without the follow-up diagnosis $d_t$ being part of the condition $c$. 
Here, we use a new SBDM model including $d_t$ in the training to create factual and counterfactual disease trajectories shown in \Cref{fig:trajectories}. 
Based on three individuals from the ADNI test set, we simulate a stable MCI scenario and a (counter-)factual progression to AD. The forecasting begins from the baseline state, and while the trajectories are generated through independent predictions for each follow-up time point, they remain both plausible and consistent with the expected course of disease progression. The predicted trajectories remain close to the measured ground truth under the actual diagnostic courses; when the target diagnosis is switched, the resulting trajectories show a marked drop in CTh, illustrating the model’s sensitivity to the imposed diagnostic changes. Although counterfactual scenarios generally lack a definitive ground truth, making it difficult to assess their realism, the simulations presented here align well with alterations in the cerebral cortical sheet observed in longitudinal studies of individuals with mild cognitive impairment (MCI) and Alzheimer's disease~\cite{Risacher2010longitudinalatrophy}.

\section{Conclusion}
In summary, we presented SBDM, the first Brownian bridge diffusion model for conditional vertex-level predictions on spherical surfaces. It consists of a new conditional denoising model, CoS-UNet, which leverages spherical convolutions and cross-attention to integrate non-Euclidean meshes and tabular conditions. We applied SBDM to the challenging task of forecasting individual cortical thickness trajectories and demonstrated significant improvements over previous approaches. The high fidelity, even on external study data, promotes SBDM's potential for straightforward application without re-training. SBDM also supports conditioning on target diagnoses to generate both factual and counterfactual trajectories, enabling novel “what-if” scenario analyses. These capabilities enable a more nuanced exploration of disease progression, offering valuable insights for personalized treatment planning and decision support.

\begin{credits}
\subsubsection{\ackname} This research was partially supported by the German Research Foundation (DFG, No. 460880779).

\subsubsection{\discintname}
The authors have no competing interests to declare that are
relevant to the content of this article.

\end{credits}

\bibliographystyle{splncs04}
\bibliography{bibliography}

\end{document}